\documentclass{article}
\usepackage{spconf,amssymb,amsmath,graphicx,color}

\definecolor{darkgreen}{RGB}{0,128,0}

\definecolor{orange}{RGB}{200,100,0}

\title{Classification of MRI data using Deep Learning and Gaussian Process-based Model Selection}
 \name{Hadrien Bertrand$^{\star \dagger}$ \qquad Matthieu Perrot$^{\dagger}$ \qquad Roberto Ardon$^{\dagger}$ \qquad Isabelle Bloch$^{\star}$}

 \address{$^{\star}$ LTCI, T\'el\'ecom ParisTech, Universit\'e Paris-Saclay, Paris, France\\
     $^{\dagger}$Philips Research, France} % TODO find correct notations
\begin{document}
\ninept
\maketitle
\begin{abstract}

The classification of MRI images according to the anatomical field of view
%in broad anatomical regions 
is a necessary task to solve when faced with the increasing quantity of medical images. In parallel, advances in deep learning makes it a suitable tool for computer vision problems. Using a common architecture (such as AlexNet) provides quite good results, but not sufficient for clinical use. Improving the model is not an easy task, due to the large number of hyper-parameters governing both the architecture and the training of the network, and to the limited understanding of their relevance. Since an exhaustive search is not tractable, we propose to optimize the network first by random  search, and then by an adaptive search based on Gaussian Processes and Probability of Improvement. Applying this method on a large and varied MRI dataset, we show a substantial improvement between the baseline network and the final one (up to 20\% for the most difficult classes).

\end{abstract}

\section{Introduction}
\label{sec:intro}

In daily clinical practice, automatic analysis of medical images so as to determine the observed anatomy would produce significant benefits in terms of time and cost, given the large number of images acquired each day. This anatomical knowledge can 1) produce reliable search tools on ever increasing datasets (currently based on manually defined tags), supporting  follow up of a specific patient anatomy or finding similar anatomies or pathologies; 2) automate the production of reports on findings (e.g. transcribing tumors location); 3) automatically enrich exams to render an augmented visualization to ease communication between clinicians and patients.

While the imaging modality (CT, MR, US, etc.) is reliably provided by the acquisition systems, the information on the imaged anatomical region (chest, abdomen, spine, etc.) is given by manual annotations. Under the pressure of maximizing equipment exploitation, this information is very prone to errors. Yet, for any other further anatomical analysis, the reliability of this information is crucial. Thus, in the present work, we aim to automatically determine the imaged anatomical region from the image content. Solving this problem within the whole variability of medical imaging (modalities, protocols, patient anatomy, pathologies, etc.) would require a dataset currently inaccessible. Thus, we choose to restrict our study to MRI modality which already encompasses many aspects of the variability of the initial problem (high variability of protocols, anatomies, etc.). Moreover, our target space is limited to four anatomical regions: spine, head, abdomen and pelvis.

The recent surge of popularity and successes of deep learning for computer vision tasks has led to a wealth of applications, including in medical imaging \cite{tmi_2016}. The commonly adopted approach for classification tasks is to retrain a popular network model (AlexNet~\cite{krizhevsky_imagenet_2012}, VGG~\cite{simonyan_very_2014}, etc.) on a specific dataset (sometimes just retraining the last few layers).
While this strategy provides often correct results on small and controlled datasets, it reveals insufficient when confronted to the high variability of actual clinical data. This is the case of our dataset, which is only built from such day to day clinical images. 
Given the high number of degrees of freedom and their correlated impact on network model performances, handcrafting better models often proves very time consuming and inefficient.

In this paper, starting from a baseline network (Section~\ref{ssec:baseline}), we present an automatic strategy to find better architectures given our clinical context. We constrain our search to a space of networks represented by a specific parametrization (Section~\ref{ssec:param}) that includes enough diversity and, at the same time, promising models (including AlexNet-like, VGG-like). Inspired by the work in~\cite{jones_taxonomy_2001}, we devise an automatic optimization process (Section~\ref{ssec:search}) to produce an ensemble of successful models (Section~\ref{ssec:ensemble}). The interest and efficiency of this strategy are demonstrated on our application in Section~\ref{sec:results}.

\section{Method}
\label{sec:method}
Under the condition that each MRI volume can be classified as one of our targeted anatomies (separating a volume into several ones if it covers several parts of the anatomy), and since MRI data are acquired in a slice by slice approach, we reduce the anatomical region prediction problem to a two dimension classification task (into abdomen, pelvis, head and spine). Assuming that information contained in a slice is rich enough, we effectively augment our dataset size. At the cost of loosing 3D information, we tend to a standard 2D-image classification problem, a good candidate for Deep Learning standard techniques.
%More formally, we aim at training a classifier $f$ which predicts anatomical regions $y_i$ from a given slice $x_i$: $y_i=f(x_i; W)$ where $W$ denotes network weights. Classes at the volume level are then obtained from a majority vote across slices.

\subsection{Handcrafted Baseline Architecture}
\label{ssec:baseline}

Several successful deep learning experiments in medical imaging are  observed using architectures inspired by AlexNet~\cite{krizhevsky_imagenet_2012} or VGG~\cite{simonyan_very_2014} (refer in particular to this special issue of \textsc{TMI}: \cite{tmi_2016}). Following these steps, after tedious parameter tuning, we converged to a model presenting a quite satisfactory behavior on our problem, but, at the same time, very difficult to further improve. 
%FIXME: parler du batch norm
Using the standard terms used by the deep learning community \cite{Goodfellow-et-al-2016-Book}, this baseline model consists of 5 blocks, each comprising a convolution layer of 64 filters of size $3\times 3$, followed by a rectified linear unit (ReLu) activation function and a max-pooling layer. The network ends with 2 fully-connected layers (resp. 4096 and 1024 units) interleaved with ReLU activations and terminated with a softmax decision layer. This network was trained by minimizing the categorical cross-entropy loss weighted by class frequency (denoted $L$ in the rest of this paper), using stochastic gradient descent (SGD) with Nesterov momentum ($m = 0.9$) and decay ($d = 10^{-6}$) .

\subsection{Parametric Architecture and Hyper-parameters}
\label{ssec:param}

\begin{figure}[htb]
\centering
\begin{minipage}[b]{0.95\linewidth}
\centering
\includegraphics[width=\linewidth]{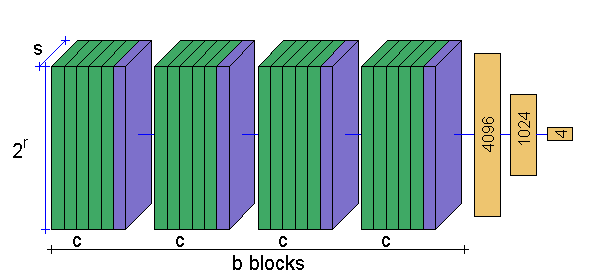}
\end{minipage}
\caption{Explored architectures}
\label{fig:par_archi}
\end{figure}

%It is likely that variations around our baseline could give rise to some improvements. 
%Thus, 
By relaxing some structural parts of our baseline architecture we define a rich family of models. This parametric family has the following structure (see Figure \ref{fig:par_archi}): (1) a convolution block comprising $b$ sections, each including $c$ convolution layers of $2^r$ filters of size $s \times s$ interleaved with ReLU activations and terminated by a max-pooling layer, (2) the fully-connected layers as in our baseline architecture, and (3) a final softmax decision layer. 
Changes within this parametric space of models may drastically transform the optimization landscape, requiring to adjust training setting accordingly (in our case: learning rate, batch size and number of epochs, all other settings remaining identical). Moreover, using or not data augmentation is also considered, since more complex models require an increased amount of information.
These architecture parameters and training settings form a collection of model hyper-parameters. Their respective ranges, detailed in Table~\ref{table:hyper}, were defined so as to fulfill memory (less than 12GB) and time constraints (training should last less than one day).
To fix ideas, each set of hyper-parameters $\Theta$ defines a classification problem that we aim to solve by training a classifier network $f$ with weights $W$, in order to predict anatomical regions $y_i$ from image slice content $x_i$: $y_i=f(x_i;W,\Theta)$.

\begin{table}
	\centering
	\begin{tabular}{ | l | c | r | }
		\hline
		Name & Range & Baseline \\ \hline
		\# blocks & $b \in [1 ; 5]$ & $5$ \\
		\# conv. layers per block & $c \in [1 ; 7]$ & $1$ \\
		\# filters per conv. layer & $2^{r}, r \in [2, 7]$ & $64$ \\
		Size of the filters & $ s \in \{3 ; 5\}$ & $3$ \\
		Learning Rate & $10^{l}, l \in [-7 ; 0]$ & $0.001$ \\
		Batch Size & $2^{a}, a \in [2 ; 8]$ & 8 \\
		\# epochs & $10e, e \in [1 ; 10]$ & $70$ \\
        Data augmentation & $g \in \{\text{Yes}, \text{No}\}$ & Yes \\
		\hline
	\end{tabular}
	\caption{Hyper-parameters list. See Section~\ref{ssec:param} for details.}
	\label{table:hyper}
\end{table}

\subsection{Hyper-parameters Optimization}
\label{ssec:search}
At this point, architecture and training parameters (a.k.a. model hyper-parameters) could be optimized with any suitable method. Given that the considered model family is exceedingly huge (more than 400K models) and that training of one network can take up to one day, any exhaustive coverage (like grid search) is intractable. Moreover, {\it a priori} discarding hyper-parameters with relative small impact remains difficult since it strongly depends on the specificity of the considered dataset~\cite{bergstra_random_2012}. Inspired by the work in~\cite{jones_taxonomy_2001} and~\cite{snoek_2012}, we developed an optimization strategy that consists of a random search warm-up stage followed by a Gaussian Process-guided search.

\subsubsection{Random Search}
\label{sssec:random}
 
Given the extent of the search space, some hyper-parameters or their combinations are expected to present only a moderate impact on our objective (minimizing slice-wise classification errors). Moreover, interesting hyper-parameter values are often consistent marginally (when integrating over a subset of other hyper-parameters).
In this respect, uniform random exploration (Random Search~\cite{Anderson_1953}) is by far more efficient than any grid search.

\subsubsection{Adaptive Search using a Gaussian Process}
\label{sssec:gp}

At some point, the coverage is sufficient to exploit the regularities of our hyper-parameters space with more advanced optimization techniques so as to accelerate the search. With such assumptions, we can consider to guide the exploration by estimating the performance of any unknown combination of hyper-parameters given performances related to all previously explored hyper-parameters values.

To this purpose, we use a Gaussian Process \cite{Rasmussen_2005} to regress performances from hyper-parameters values. Smoothness is easily enforced thanks to a Gaussian Kernel whose scale is optimized by maximizing log-marginal-likelihood for each hyper-parameter. Estimation is very efficient; thus, the whole hyper-parameters space can be entirely covered to get, at any point $\Theta$, an estimated loss following a Gaussian model: $\mathcal{N}(\hat{L}\left(\Theta\right), \hat{\sigma}^2\left(\Theta\right))$.

% The Gaussian Process can be used to determine the most relevant parameters using a method called Automatic Relevance Determination (ARD~\cite{bishop_pattern}). The idea is that each hyper-parameter has its own kernel, and so its own length scale. The optimum is the one that maximizes the marginal likelihood. The relevance is defined as its inverse. 

Many optimization strategies can be considered atop of such probabilistic estimations. Practically, the next hyper-parameters proposal is the one that maximizes the probability $PI$ of overcoming a given target $L^*$~\cite{jones_taxonomy_2001}:
\begin{equation}
PI(\Theta) = \Phi \left( \frac{L^{*} - \hat{L}(\Theta)}{\sigma(\Theta)} \right)
\end{equation} where $\Phi$ denotes the normal cumulative distribution function. To foster a good balance between optimizing locally already identified good proposals and exploring farther regions with potential improvements, the optimization is driven by two targets in parallel: the best loss seen so far and an improvement of 25\% from it.

\subsection{Ensemble model}
\label{ssec:ensemble}

\begin{figure*}[bt]
	\begin{minipage}[b]{.99\linewidth}
		\centering
		\centerline{\includegraphics[width=18cm]{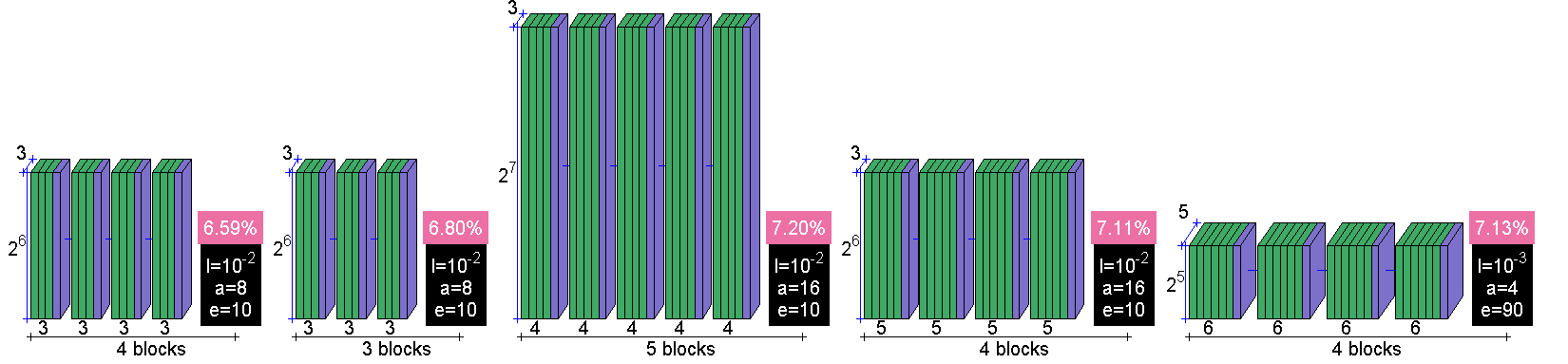}}
	\end{minipage}
	\caption{Architecture of the 5 best models. Height represents the number of filters per layer, depth the size of the filters.}
	\label{fig:best_models_archi}
\end{figure*}

At the end, our optimization process yields a collection of models ranked by their performance. The quality of this assessment is naturally limited since the size of the validation set used in this respect cannot encompass the diversity of clinical reality. Cross-validation could be used to get a better estimator of the performance, but we cannot practically afford its costs. Nevertheless, best models present very similar error rates and, at the same time, a good diversity of architectures (see Figure \ref{fig:best_models_archi}) that can be leveraged. In this paper, we select the top-ten models and build a robust classifier by averaging their predictions.

\section{Results}
\label{sec:results}

\subsection{Dataset}
\label{ssec:dataset}

\begin{figure}[ht]
	\begin{minipage}[b]{.24\linewidth}
		\centering
		\centerline{\includegraphics[width=2.0cm]{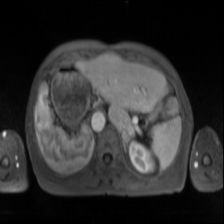}}
		%\vspace{0.2cm}
	\end{minipage}
	%\hfill
	\begin{minipage}[b]{.24\linewidth}
		\centering
		\centerline{\includegraphics[width=2.0cm]{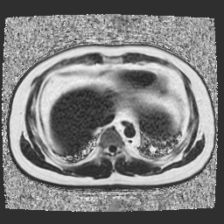}}
	%	\vspace{0.2cm}
	\end{minipage}
	\begin{minipage}[b]{.24\linewidth}
		\centering
		\centerline{\includegraphics[width=2.0cm]{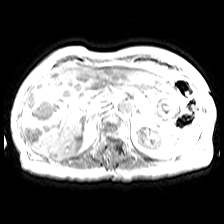}}
	\end{minipage}
	\hfill
	\begin{minipage}[b]{.24\linewidth}
		\centering
		\centerline{\includegraphics[width=2.0cm]{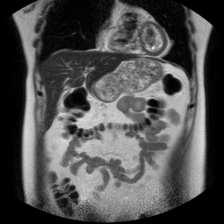}}
	\end{minipage}
	\caption{A selection of axial and coronal abdomen slices showing the diversity and the complexity of our dataset.}
	\label{fig:abdomen}
\end{figure}

The dataset consists of MRI images coming from a variety of hospitals and machines across the world (such as the \textit{Centre Hospitalier Lyon-Sud, France} or \textit{ Johns Hopkins University, USA}). As a consequence our images display a large variety of protocols (see Figure~\ref{fig:abdomen}) as well as resolution and number of slices per volume. In this paper, considered regions are limited to: abdomen, head, pelvis and spine (table~\ref{table:dataset} sums up the content of our dataset).

Our dataset is splitted in a training set for the optimization of the weights $W$, a validation set for model selection (optimization w.r.t hyper-parameters) and a test set for model evaluation (resp. $50 \%$, $25 \%$, $25 \%$). The separation is done volume-wise to take into account intra-subject slices correlations. Volumes containing multiple classes are split by anatomical regions and can end up in different sets. This raises the difficulty of the task since, in case of overfitting, predictions will be wrong at validation or testing phases.
% FIXME: pour la version finale: combien de volumes multi-classes ?
We also stratified classes across sets, giving us a proportion of slices per class close to the proportion of volumes per class.
% balanced the sets so that each class is represented with the same proportion in each set

Finally, each slice is subject to a unique step of preprocessing: it is resized to $64 \times 64$ pixels, a good trade-off between time constraints and quality of information.

Data augmentation consists in generating 80 000 images per epoch, which is 4 times as many images as the training set. The augmentation is done by applying translations, shearing and rotations, zooming, and adding noise.

\begin{table}
	\centering
	\begin{tabular}{ | l | c | r | }
		\hline
		Body Part & \# Volumes & \# Slices \\ \hline
		Abdomen & 282 & 11532 \\
		Head & 301 & 9032 \\
		Pelvis & 225 & 8854 \\
		Spine & 386 & 7732 \\
		\hline
	\end{tabular}
	\caption{Content of the dataset.}
	\label{table:dataset}
\end{table}
% FIXME: pour la version finale: ajouter min/mean/max slices par volumes

\subsection{Hyper-parameters Optimization}
\label{ssec:optim}

\begin{figure}[htb]
	\begin{minipage}[b]{.5\linewidth}
		\centering
		\centerline{\includegraphics[width=4.2cm]{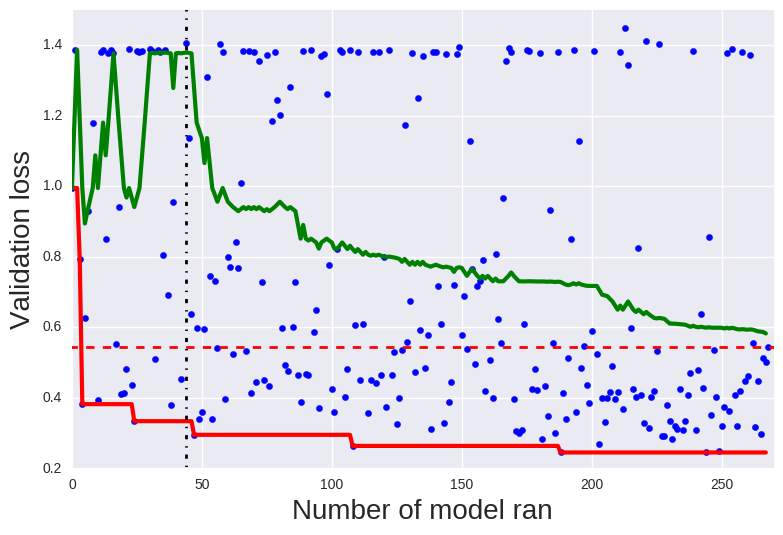}}
	\end{minipage}
	\begin{minipage}[b]{.5\linewidth}
		\centering
		\centerline{\includegraphics[width=4.2cm]{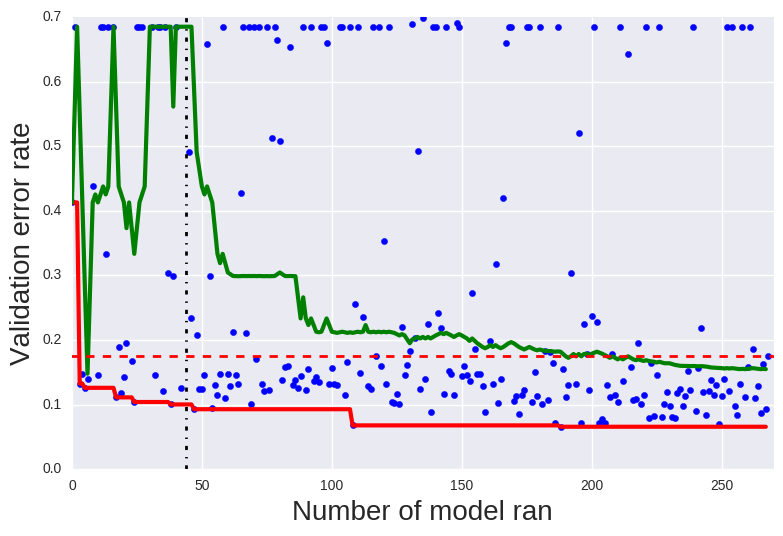}}
	\end{minipage}
	\caption{Loss (left) and error rates (right) on validation dataset in function of the number of models considered so far by the optimization process. In green: running median, in solid red: running min, in dashed red: baseline performance, in dashed black: iteration from which random search is stopped in favor of adaptive search.}
	\label{fig:loss_acc_vs_iterations}
\end{figure}

The hyper-parameters were optimized in two steps, 47 iterations of random search followed by an adaptive search (as described in Section~\ref{sssec:gp}). The entire process is depicted in Figure~\ref{fig:loss_acc_vs_iterations}. Adaptive search presents quickly an important increase of the proportion of models with good performance (supported by the decreasing running median of the loss). Thus, selected combinations are on average better than random search. 

Many proposals present both a high loss and a high error rate. These corresponds to models that put all images in the same class, in this case: abdomen, which accounts for around $30 \%$ of the dataset (implying $70 \%$ of error rate).

\subsection{Test accuracy}
\label{ssec:accuracy}

\begin{figure}[ht]
	\begin{minipage}[b]{.9\linewidth}
		\centering
		\centerline{\includegraphics[width=8.0cm]{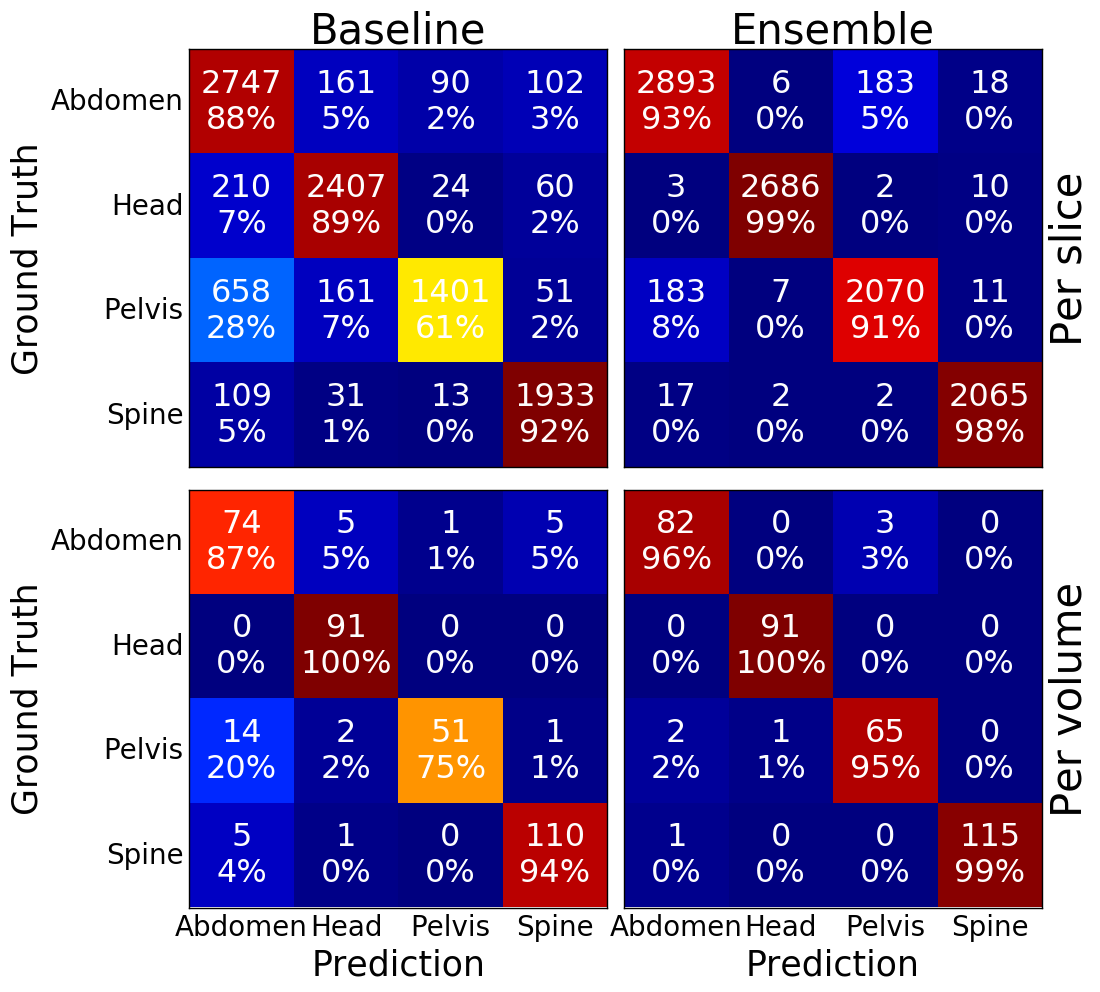}}
	\end{minipage}
	\caption{Confusion matrices on the test set, for the baseline network (left) and the ensemble (right), computed on slices (first row) and on volumes (second row).}
	\label{fig:conf}
\end{figure}

Figure \ref{fig:best_models_archi} shows the architecture of some of the best models chosen for our ensemble. Despite the first two differing only in the number of blocks, others display variations across all hyper-parameters except data augmentation, which is always turned on. The learning rate is in a small range, either $0.0001$ or $0.001$, and the batch size is small (less than 16). Those networks tend to be deep (min. 8 convolutional layers) and the other hyper-parameters use a wide range of values.

In terms of accuracy, the ensemble is slightly better than the best model alone, however the ensemble benefits from a reduced bias.

Figure~\ref{fig:conf} shows the confusion matrices on the test set of the baseline and the ensemble of the 10 best models, demonstrating a substantial improvement on the classification of all anatomical regions. Most of the errors come from pelvis and abdomen, which was expected since the delimitation between those regions is ill-defined. In both cases pelvis is the class with the highest error.

For the volume classification, the choice of class is done by a majority vote on all slices of the volume. This gives us a higher accuracy. The ensemble misclassifies 444 slices from 71 volumes, but only 7 volumes produce errors. The misclassified slices usually correspond to the first or last one of a volume, containing little information or being nearly part of another anatomical region.

\subsection{Slice by slice analysis of a volume}
\label{ssec:volume}

\begin{figure}[ht]
	\begin{minipage}[b]{.99\linewidth}
		\centering
		\centerline{\includegraphics[width=8.0cm]{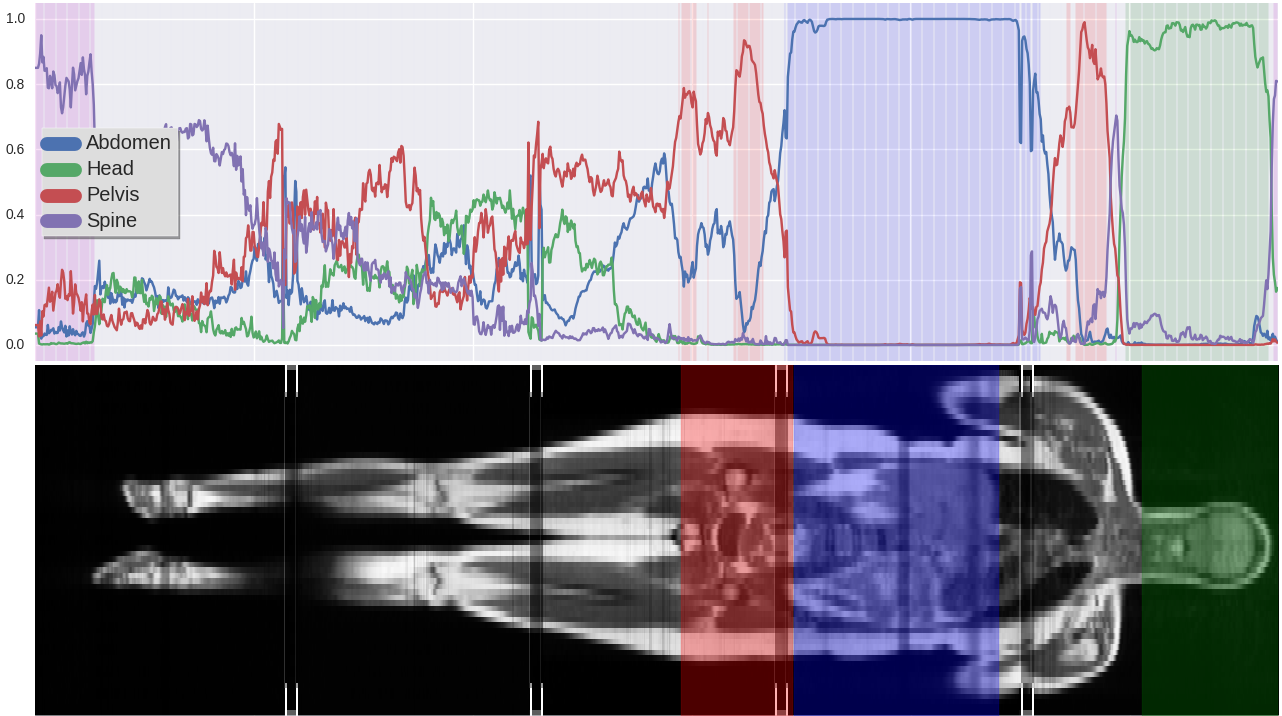}}
	\end{minipage}
	\caption{Slice by slice classification on a full body volume. Top: Class probabilities. Filled areas correspond to decision made when the probability is higher than 0.7. Bottom: Volume and ground truth.}
	\label{fig:full_body}
\end{figure}

As an interesting example, we analyzed a full body volume by classifying each of its slices through our ensemble model. For each slice, the predicted class is the one with a probability higher than 0.7, and if no class meets this criterion, then we do not choose any. As we can see in Figure~\ref{fig:full_body}, the network is doing well at identifying the abdomen and the head. It also identifies correctly the pelvis, with some uncertainty. No class is dominant for most of the legs, however the feet are considered as spine with high probability. It also mistakenly identifies the neck as pelvis.

Those mistakes could be corrected by using a more complex decision criterion than a simple probability cutoff. We also expect that adding more anatomical regions to our dataset will allow for a better localization of the present regions.

\section{Discussion and Conclusion}
\label{sec:discussion}

To the problem of finding more accurate networks than handcrafted ones, we have answered with two viable strategies. Random search is as easy to implement as grid search and quickly improves on the baseline, which makes it perfect for time-constrained situations. Without this constraint and at a higher cost in implementation, an adaptive search based on Gaussian Processes explores a range of highly accurate models suited for ensembling.% The search can be started immediately in adaptive mode, as the GP will have high uncertainty everywhere, resulting in an almost uniform distribution.

For the hyper-parameters where there is a ``correct" answer, such as the learning rate (0.001) and the presence of data augmentation, the guided search quickly converges and most models inherit their values. We should remove them on further analysis and instead explore a wider range of architectures by adding hyper-parameters controlling the fully-connected layers, such as the number of units, the number of layers, adding new types of layers such as dropout and batch normalization placed across the network or even explore other learning method such as RMSProp or Adadelta. 

One limitation of the current system is that some hyper-parameters depend on the value of others. For example we are unable to choose a different number of filters per layer as the number of layers is not fixed. We also limited the range of some hyper-parameters such as the filter size so as to have networks that would fit in memory and be trained in a reasonable amount of time. Only a subset of values combinations would cause a failure. Further work could incorporate constraints on time and memory either by precise estimations when possible or by measures during training to produce estimations with a GP. 

Results on volume classification were very satisfactory. Since it is done at slice level, we have obtained a decent localization tool which shows the robustness of our ensemble. Further work will focus on adding more anatomical regions, which might require splitting slices in patches to identify smaller regions such as organs.

\label{sec:ref}
\bibliographystyle{IEEEbib}
\bibliography{refs}

% To start a new column (but not a new page) and help balance the last-page
% column length use \vfill\pagebreak.
% -------------------------------------------------------------------------
\vfill
\pagebreak

\end{document}